\documentclass[sigconf,nonacm]{acmart}
\renewcommand\footnotetextcopyrightpermission[1]{} 
\usepackage{subcaption}
\usepackage{tikz}
\usetikzlibrary {calc,bayesnet,shapes,backgrounds,arrows.meta}
\setcopyright{none}
\acmDOI{}

\acmISBN{}

\acmConference[]{}{}{}
\copyrightyear{}

\begin{document}

\tikzset{
    connector/.style={
     -latex,
     font=\scriptsize
    },
    rectangle connector/.style={
        connector,
        to path={(\tikztostart) -- ++(#1,0pt) \tikztonodes |- (\tikztotarget) },
        pos=0.5
    },
    rectangle connector/.default=-2cm,
    straight connector/.style={
        connector,
        to path=--(\tikztotarget) \tikztonodes
    }
}

\title[Exploitation of Neuroevolutionary Information: Analyzing Past for a More Efficient Future]{On the Exploitation of Neuroevolutionary Information: Analyzing the Past for a More Efficient Future}

\author{Unai Garciarena}
\authornote{Corresponding author.}
\orcid{0000-0003-2425-8340}
\affiliation{%
\institution{Department of Computer Science and Artificial Intelligence\\University of the Basque Country}   
\streetaddress{P. Manuel Lardizabal 1}
\city{Donostia-San Sebastian} 
\state{Spain} 
\postcode{20018}
}
\email{unai.garciarena@ehu.eus}

\author{Nuno Lourenço}
\orcid{0000-0002-2154-0642}
\affiliation{%
\institution{CISUC, Department of Informatics Engineering}
\streetaddress{University of Coimbra}
\city{Coimbra} 
\state{Portugal} 
}
\email{naml@dei.uc.pt}

\author{Penousal Machado}
\orcid{0000-0002-6308-6484}
\affiliation{%
\institution{CISUC, Department of Informatics Engineering}
\streetaddress{University of Coimbra}
\city{Coimbra} 
\state{Portugal} 
}
\email{machado@dei.uc.pt}

\author{Roberto Santana}
\orcid{0000-0002-1005-8535}
\affiliation{%
\institution{Department of Computer Science and Artificial Intelligence\\University of the Basque Country}   
\streetaddress{P. Manuel Lardizabal 1}
\city{Donostia-San Sebastian} 
\state{Spain} 
\postcode{20018}
}
\email{roberto.santana@ehu.eus}

\author{Alexander Mendiburu}
\orcid{0000-0002-7271-1931}
\affiliation{%
\institution{Department of Computer Architecture and Technology\\University of the Basque Country}   
\streetaddress{P. Manuel Lardizabal 1}
\city{Donostia-San Sebastian} 
\state{Spain} 
\postcode{20018}
}
\email{alexander.mendiburu@ehu.eus}


\begin{abstract}
Neuroevolutionary algorithms, automatic searches of neural network structures by means of evolutionary techniques, are computationally costly procedures. In spite of this, due to the great performance provided by the architectures which are found, these methods are widely applied. The final outcome of neuroevolutionary processes is the best structure found during the search, and the rest of the procedure is commonly omitted in the literature. However, a good amount of \textit{residual information} consisting of valuable knowledge that can be extracted is also produced during these searches. In this paper, we propose an approach that extracts this information from neuroevolutionary runs, and use it to build a metamodel that could positively impact future neural architecture searches. More specifically, by inspecting the best structures found during neuroevolutionary searches of generative adversarial networks with varying characteristics (e.g., based on dense or convolutional layers), we propose a Bayesian network-based model which can be used to either find strong neural structures right away, conveniently initialize different structural searches for different problems, or help future optimization of structures of any type to keep finding increasingly better structures where uninformed methods get stuck into local optima.
\end{abstract}

%
%


\keywords{Generative adversarial networks, evolutionary algorithms, probabilistic graphical models, neural architecture search}

\maketitle

\section{Introduction}

Since deep neural network (DNN) models were \textit{rediscovered} earlier this decade as a greatly accurate model for image classification \cite{krizhevsky_imagenet_2012}, the increase in their popularity has led them to be (successfully) applied to many other tasks, such as dealing with temporal data \cite{hochreiter_long_1997,cho_learning_2014} or generative modeling with, for example, generative adversarial networks (GAN) \citep{goodfellow_generative_2014, schmidhuber_generative_2020}. As the accessibility to augmenting computational resources (mainly graphic processing units) has broadened for researchers, increasingly complex DNN structures have been proposed in the last few years \cite{krizhevsky_imagenet_2012,szegedy_going_2015} and although top performing DNNs can still be \textit{manually} designed \cite{szegedy_rethinking_2016,szegedy_inception-v4_2017}, this technological advance has caused a drop-off in the feasibility of designing these models by hand. This, at the same time, has enabled neural architecture search (NAS) \cite{elsken_simple_2017,elsken_neural_2019,zoph_learning_2018,liang_evolutionary_2018}, a research field that encompasses methods whose main goal is to automatize this process, to gain considerable popularity. Although approaches inspired by reinforcement learning \cite{zoph_learning_2018} and local searches \cite{elsken_simple_2017} have been proposed, the usage of evolutionary algorithms (EA) has hoarded a large portion of this research field \cite{miikkulainen_evolving_2019,liang_evolutionary_2018,stanley_compositional_2007}. This has resulted in this particular application of algorithms to NAS becoming a subfield itself, known as neuroevolution (NE) \cite{stanley_designing_2019}.

EAs have proven to be effective methods when it comes to giving good solutions to complex optimization problems under a reasonable amount of time. They are commonly population-based algorithms, and their success can depend on many design choices, such as codification, search landscape, selected model or operators. In addition, for complex problems, randomly initializing the individuals that compose the first population can be a poor strategy that can lead to algorithms being unable to search for promising areas or requiring a large computational effort to locate them \cite{kazimipour_review_2014,bajer_population_2016,elsayed_sequence-based_2016}. This effect is not exclusive to EAs, as similar issues can also happen with local search approaches.

Neuroevolution falls into that complex problem category, as the large number of design options available to build a DNN provoke a combinatorial explosion. Thus, an optimization process that uses a random starting point and relies on uninformed operators is likely to require a large number of steps to approach areas with promising solutions, if capable at all of doing so. In a context in which extra information provided to the algorithm can be crucial, this paper aims at studying how to extract as much advantage as possible from evolutionary runs that have already been carried out, and possible applications for the obtained information. To that end, we employ previous diverse use cases of evolved GANs whose final goal was generating data with entirely different structures, i.e., images \cite{costa_coevolution_2019} and Pareto set approximations \cite{garciarena_analysis_2020}.

More specifically, we record the characteristics of the best GANs found during these evolutionary searches. By modeling these configurations, it is possible to determine the attributes desired in top performing GANs, which would be helpful at the time of creating new DNNs (e.g., when initializing individuals for NE algorithms or when aiming for structures that are expected to work right away).
Additionally, if the model were able to recognize GAN definitions which are likely to deliver good results when evaluated, multiple evaluations with no positive impact on future structure searches could be avoided. This would lessen the computationally costly nature of NE algorithms, which is one of their main drawbacks~\cite{white_local_2020}.

Because of their suitability to perform the tasks described above, and the strong dependencies present among the multiple components factoring in a GAN structure, we consider probabilistic graphical models (PGM) \cite{pearl_probabilistic_1988} an appropriate tool to model this data. Particularly, this work makes use of Bayesian networks (BN) \cite{castillo_expert_1997} to model the dependencies of the data. On top of the two advantages mentioned in the previous paragraph, the usage of BN would add a third one. These models produce a graph-like layout that can be visually interpreted, which could result in a better understanding of what characteristics cause a GAN to produce good quality results. In this paper, we propose a BN-based model for effectively capturing the dependencies between GAN components that can be posteriorly used for different purposes: obtaining an estimate of the quality of a known GAN structure, probabilistically generating strong-performing structures, or guiding NAS runs.

The paper is organized as follows: Section~\ref{sec:rw} contains a brief review of works that are particularly relevant for this proposal. In Section~\ref{sec:proposal}, we make a detailed introduction of the proposed algorithm. Once the proposal is presented, its validity and generalization capacities are tested in Sections~\ref{sec:mainExp} and \ref{sec:secondaryExp}. The most important conclusions are summarized in Section~\ref{sec:conclusions}.

\section{Work Background} \label{sec:rw}

In this section we introduce the three main components of out proposal: Generative Adversarial Networks, Bayesian Networks, and GAN neuroevolution and other NAS. It needs to be noted that, although the approach has only been applied to GAN structure optimization, it is transferable to other DNN definitions. 

\subsection{Generative adversarial networks} \label{subsec:usecase}

A traditional GAN is composed of two DNNs with opposing goals, a generator $G$ and a discriminator $D$. While the latter aims at being able to discern real data observations from fake ones, the former pursues fooling the discriminator by generating data samples as similar to the real ones as possible. The final objective of the GAN is to learn realistic generative models of high-dimensional data.

Depending on the type of data to be \textit{replicated}, the DNNs are based on one architecture type or another. When facing temporal data (e.g., financial stocks), GANs can be based on recurrent neural networks \cite{yoon_time-series_2019}; when data features are spatially related (as in a picture of a face), CNNs -and their counterpart, transposed CNNs- can be used as a basis for the models \cite{karras_progressive_2018}; in case no knowledge about the structural dependencies of the data is known, the default selection of MLPs \cite{cybenko_approximation_1989} for both DNNs is a fair choice.

Despite the fact that they can be based on completely different architectures, the way in which the two models within a GAN interact with each other to build a generator as accurate as possible does not vary. A generator $G$, partially defined by its parameters $\theta_G$, receives random noise sampled from a distribution ($p_z({\bf{z}}) \approx \mathcal{N}(\mu,\sigma)$ being the common choice) and produces samples in the space of the original data $\bf{x}$:  $G({\bf{z}},\theta_g) \rightarrow  \hat{\bf{x}}$.

The discriminator, using its parameters $\theta_d$, can receive input from either the set of original points ${\bf{x}}$ or the generator $\hat{\bf{x}}$ and provides  probability values of the inputs emanating from ${\bf{x}}$.

Equation~\eqref{eq:LOSSGANN}, which shows the min-max game that can be used for training GANs, captures these opposing goals

\begin{equation}
      \min_G \max_D = \mathbb{E}_{{\bf{x}} \sim p_{data}({\bf{x}})} [log(D({\bf{x}})] +   \mathbb{E}_{{\bf{z}} \sim p_{\bf{z}}({\bf{z}})} [log(1-D(G({\bf{z}}))]   \label{eq:LOSSGANN}
\end{equation}
where  $\mathbb{E}_{{\bf{x}} \sim p_{data}({\bf{x}})}$ and $\mathbb{E}_{{\bf{z}}\sim p_{\bf{z}}({\bf{z}})}$ represent the expectations with respect to the original data distribution $p_{data}$ and the latent variable distribution $p_{\bf{z}}({\bf{z}})$ respectively.

\subsection{Bayesian networks}

Bayesian networks are probabilistic graphical models that are used to represent sets of variables, together with their (in)dependencies, by means of directed acyclic graphs (DAG) \cite{pearl_causality:_2000,pearl_probabilistic_1988}. Each node of the DAG represents a variable, and the (non) existence of an arc between two nodes represents the (in)dependency between them. When the graph structure of a BN is given by an expert, only the parameters are learnt. If that is not the case, the structure must be deduced from available data via different techniques as well.

Therefore, the automatic learning of a BN consists of two phases: i) devising the topology of the network, i.e., which connections should and should not exist, and ii) learning the conditional or marginal probabilities of each variable \cite{larranaga_estimation_2001}.

In this work, we use BNs to codify the probabilistic relationship between the parameters used to construct the DNNs, e.g., activation functions, size of the layers, or weight initialization procedures. Once the BN has been learnt from data gathered during the evolution of different DNNs, we will perform inference looking for the most probable (promising) configurations, or calculating the probability given by the BN to a particular set of DNN characteristics.

\subsection{GAN neuroevolution and other NAS} \label{subsec:evo_intro}

COEGAN \cite{costa_coevolution_2019} proposes a coevolutionary NE approach based on DeepNEAT \cite{miikkulainen_evolving_2019} for evolving GANs. The authors test the proposed methodology by employing it on different image generation problems, and because of this, the structures of both generators and discriminators are based on CNNs. The indirect encoding employed in COEGAN uses transformation maps to implement the components encoded in the genotype to a sequence of layers, the phenotype. These layers can be of one of three possible types: linear, convolution (exclusive to the discriminator) and transposed convolution (exclusive to the generator). COEGAN evolves discriminators and generators in two different populations, each of which applies a speciation mechanism inspired by the NEAT \cite{stanley_evolving_2002} selection mechanism in order to promote innovation in each subpopulation. This way, the survival of \textit{new} individuals until maturity is promoted. This approach does not consider the evolution of DNN parameters, which are trained using a gradient descent method. 

In order to explore the search space, COEGAN relies on two mutation operators which are randomly applied to a given individual: add layer, and delete layer. These operators are applied to a random DNN in a GAN, and all of them involve complete layers (a layer can be added to either the discriminator or a generator, it can be dense or transposed convolutional/convolutional depending on the network, and includes an activation function).

To quantify the quality of the GANs being evolved, the authors use the Frèchet inverted distance (FID) \cite{heusel_gans_2017} over the samples of the GANs after they are trained with the MNIST database \cite{lecun_mnist_2010}.

The work in \cite{garciarena_analysis_2020} proposes the evolution of GANs based on MLPs, although the NE approach shares some conceptual characteristics with the one presented above, namely, the encoding of the DNNs (weights are not evolved and the DNNs are specified by lists of layers, which are codified by their size and activation function). In this case, the DNNs are not evolved isolated from each other, each individual being composed of a fully functioning GAN with two DNNs. The variation operators also share some common factors: the mutation operators include the addition, the deletion and the change of a layer (although this last operator can also vary parts of a layer on top of the whole layer altogether). A crossover operator which interchanges the two DNNs in a GAN is also defined, which enables interaction between different sections of individuals.

Within the NE framework, the capacities of the GANs are gauged by measuring the inverted generational distance (IGD) \citep{coello_coello_study_2004} between a set of points known to be in the Pareto set of a bi-objective problem and the points generated by the GAN, after being trained to generate points from that set. As a reference for the ground truth, the benchmark described in \cite{li_multiobjective_2008} was used, which provides difficult optimization problems along with their corresponding solutions. From the 9 functions available in this benchmark, 8 were used. Additionally, taking advantage of the fact that the benchmark allows an arbitrary number of variables in the optimization suites, two different solution sizes were investigated, 10 and 784.

The quality of the evolved GANs was further validated by a successful application of the evolved structures to the 2D 8-Gaussian approximation problem \cite{metz_unrolled_2016}. 

E-GAN, an approach which exploits the antagonistic nature of GANs in a different manner when compared to the previous two approaches, was proposed in \cite{wang_evolutionary_2019}. In this example, only generators are evolved, by making them adapt to the environment, which consists of the discriminator. In this evolutionary formulation, the environment is dynamic in the sense that each time a generator is trained so is the \textit{global} discriminator. That way, it can be assumed that the discriminator can offer a strong performance, due to the intense training it receives. This enables the procedure to focus its operators on the generators, which is the final outcome of a GAN. 
The variation operators are another novel proposal in this work. Instead of being designed for exploring the structural space of the generator (which remains fixed during the whole procedure), they consist of the application of one training epoch, making use of one of three available loss functions. Therefore, an individual poised to advance to the next population is subjected to one of three mutation operators which also (ideally) improve the ability of the discriminator to distinguish real from fake samples.

All the evolutionary proposals mentioned in this section start from randomly initialized individuals which, although it allows a wide exploration of the search space, results in several very costly evaluations until an acceptable convergence point is reached. By adjusting the initial individuals to patterns known to be beneficial in the design of GANs, they could be largely more efficient. Furthermore, it is theorized that constraining the NE exploration to reduced areas can produce results as good (if not better) than not doing so \cite{white_local_2020}. Additionally, All GANs are evaluated, which not only includes the training of the structures, but also computation of the fitness function. This can also result in significant increases in the total elapsed time.

Recently, some local-search oriented NAS algorithms have also been proposed. Despite their \textit{limited} search scope, the quality of the results they produce have made them a viable alternative to the more costly NE approaches. The work in \cite{elsken_simple_2017} takes advantage of the network morphism (NM) framework defined in \cite{wei_network_2016} and uses it as a tool for a Neural Architecture Search by Hillclimbing (NASH). It uses a simple structure as a starting point (the teacher network), and uses the NM operators to widen and deepen the model, generating candidate neighbors (student networks, which inherits the weights of the teacher), which are accepted or rejected as the new teacher depending on their performance relative to the previous teacher model. Once the NASH algorithm reaches a convergence point, the resultant structure is also trained starting from random weights in order to test the validity of the weight inheritance concept. The results show the efficiency gains found in using the NASH approach rather than training a complete DNN from scratch, as better results are obtained faster. Finally, the authors compare their approach to other structure search methods in the literature, obtaining competitive results in much more reduced computing time for the CIFAR-10 and CIFAR-100 databases.

\section{BN-assisted NAS} \label{sec:proposal}

The fact that there is no consensus on the best neural architecture -nor neural cell architecture- for any given problem, implies the existence of dependencies between the different components in neural models. This is especially applicable to GANs, as they are commonly composed of two DNNs, which supposes double the parameters. More specifically, the DNNs in a GAN are characterized by how frequently they are trained, the number of layers of each DNN and their characteristics (e.g., their activation and weight initialization functions, their number of neurons in MLPs, the number of filters and their sizes and strides for CNNs, etc.), as well as other \textit{global} parameters, such as the distribution followed by the noise fed to the discriminator or the loss function of the model.

\subsection{Metamodel choice} \label{sec:design}

The model used to represent the characteristics of good quality GANs must be able to capture the dependencies underlying in their structures. PGMs is one such model family, and they are the model of choice for this work for three main reasons.

Firstly, they are naturally modular so that complex dependency structures can be described and handled by a careful combination of simple elements. Secondly, they are visually representative, which would help to understand the decision making process underlying the final product we seek, the sample \cite{lauritzen_graphical_1996}. Finally, there are different ways of introducing \textit{expert knowledge} into the model. More specifically, the characteristics of the chosen GANs are going to be modeled using Bayesian networks (BN) \cite{castillo_expert_1997}.

\begin{figure}
  \begin{center}
    \begin{tikzpicture}[scale=1, transform shape]
      \newcounter{varsep}
      \setcounter{varsep}{8}
      \newcounter{bnsep}
      \setcounter{bnsep}{55}
      \newcounter{netsep}
      \setcounter{netsep}{30}
      \newcounter{subsep}
      \setcounter{subsep}{50}
      \newcounter{xxsep}
      \setcounter{xxsep}{10}
      \newcounter{xysep}
      \setcounter{xysep}{20}
      
      \node at (-0.5, 6.3) (Metamodel) {\Large\textbf{Metamodel}};
      \node at (-0.4, 2) (Supermodel) {\textbf{Supermodel}};
      \node at (-0.4, 1.2) [rectangle, fill=blue!20,draw=blue] (supervars) {$d_g$, $d_d$};

      \node at (3.6, 6) (submodels) {\textbf{Submodels}};
      
      \node at (3, 5) (bn0) {\textbf{$\text{BN}_0$}};
      \path let \p1 = (bn0) in node  at (\x1+\value{netsep},\y1) [circle, draw,inner sep=1pt] (bn0n0) {$x_1$};
      \path let \p1 = (bn0n0) in node  at (\x1+\value{xxsep},\y1-\value{xysep}) [circle, draw,inner sep=1pt] (bn0n1) {$x_2$};
     
      \path let \p1 = (bn0) in node  at (\x1,\y1-\value{subsep}) (bn1) {\textbf{$\text{BN}_1$}};
      \path let \p1 = (bn1) in node  at (\x1+\value{netsep},\y1) (bn1n0) [circle, draw,inner sep=1pt] {$x1$};
      \path let \p1 = (bn1n0) in node  at (\x1+\value{xxsep},\y1-\value{xysep}) [circle, draw,inner sep=1pt] (bn1n1) {$x_2$};
      \path let \p1 = (bn1n0) in node  at (\x1-\value{xxsep},\y1-\value{xysep}) [circle, draw,inner sep=1pt] (bn1n2) {$x_3$};

      \path let \p1 = (bn1) in node  at (\x1,\y1-\value{subsep}) (bn2) {\textbf{$\text{BN}_2$}};

      \path let \p1 = (bn2) in node  at (\x1+\value{netsep},\y1) (bn2n0) [circle, draw,inner sep=1pt] {$x1$};
      \path let \p1 = (bn2n0) in node  at (\x1+\value{xxsep},\y1-\value{xysep}) [circle, draw,inner sep=1pt] (bn2n1) {$x_2$};
      \path let \p1 = (bn2n0) in node  at (\x1-\value{xxsep},\y1-\value{xysep}) [circle, draw,inner sep=1pt] (bn2n2) {$x_3$};
      \path let \p1 = (bn2n0) in node  at (\x1,\y1-2*\value{xysep}) [circle, draw,inner sep=1pt] (bn2n3) {$x_4$};

      \path let \p1 = (bn2) in node  at (\x1+5,\y1-2*\value{subsep}) (bn3) {\textbf{$\text{BN}_{m\times l}$}};
      \path let \p1 = (bn3) in node  at (\x1+\value{netsep}-5,\y1) (bn3n0) [circle, draw,inner sep=1pt] {$x1$};
      \path let \p1 = (bn3n0) in node  at (\x1+\value{xxsep},\y1-\value{xysep}) [circle, draw,inner sep=1pt] (bn3n1) {$x_2$};
      \path let \p1 = (bn3n0) in node  at (\x1-\value{xxsep},\y1-\value{xysep}) [circle, draw,inner sep=1pt] (bn3n2) {$x_3$};
      \path let \p1 = (bn3n0) in node  at (\x1,\y1-2*\value{xysep}) [circle, draw,inner sep=1pt] (bn3n3) {$x_4$};

      \begin{scope}[on background layer]
        \node[fit= (Supermodel)(supervars)(submodels)(bn3n3)(bn3n1),rectangle,fill=red!30,draw=red,fill opacity=0.5, minimum width = 7cm, minimum height = 11cm] (components){};

        \node[fit= (bn0)(bn0n1),rectangle,shape border rotate=270,fill=blue!20,draw=blue] (bn0s){};
        \node[fit= (bn1)(bn1n1),rectangle,shape border rotate=180,fill=blue!20,draw=blue] (bn1s){};
        \node[fit= (bn2)(bn2n1)(bn2n3),rectangle,shape border rotate=180,fill=blue!20,draw=blue] (bn2s){};
        \node[fit= (bn3)(bn3n1)(bn3n3),rectangle,shape border rotate=180,fill=blue!20,draw=blue] (bn3s){};
        
        \node[fit= (submodels)(bn3s),rectangle,shape border rotate=180,draw, minimum width = 2.8cm, minimum height = 10.4cm] (subms){};
        
        \draw[-Latex] (bn0n0)--(bn0n1) node[right]{};
        
        \draw[-Latex] (bn1n1)--(bn1n0) node[right]{};
        \draw[-Latex] (bn1n1)--(bn1n2) node[right]{};
        
        \draw[-Latex] (bn2n0)--(bn2n1) node[right]{};
        \draw[-Latex] (bn2n0)--(bn2n2) node[right]{};
        \draw[-Latex] (bn2n1)--(bn2n3) node[right]{};
        
        \draw[-Latex] (bn3n2)--(bn3n1) node[right]{};
        \draw[-Latex] (bn3n2)--(bn3n3) node[right]{};
        \draw[-Latex] (bn3n1)--(bn3n3) node[right]{};
        \draw[-Latex] (bn3n0)--(bn3n1) node[right]{};
        
      \end{scope}
      
      \path let \p1 = (bn0s) in node  at (\x1-\value{bnsep},\y1+\value{varsep}) (vars00) {$d_d=1$};
      \path let \p1 = (bn0s) in node  at (\x1-\value{bnsep},\y1-\value{varsep}) (vars01) {$d_g=1$};
      
      \path let \p1 = (bn1s) in node  at (\x1-\value{bnsep},\y1-\value{varsep}) (vars10) {$d_g=1$};
      \path let \p1 = (bn1s) in node  at (\x1-\value{bnsep},\y1+\value{varsep}) (vars11) {$d_d=2$};
      
      \path let \p1 = (bn2s) in node  at (\x1-\value{bnsep},\y1-\value{varsep}) (vars20) {$d_g=2$};
      \path let \p1 = (bn2s) in node  at (\x1-\value{bnsep},\y1+\value{varsep}) (vars21) {$d_d=2$};
      
      \path let \p1 = (bn3s) in node  at (\x1-\value{bnsep},\y1-\value{varsep}) (vars30) {$d_g=m$};
      \path let \p1 = (bn3s) in node  at (\x1-\value{bnsep},\y1+\value{varsep}) (vars31) {$d_d=l$};
      
      \node[fit=(vars00)(vars01)] (vars0){};
      \node[fit=(vars10)(vars11)] (vars1){};
      \node[fit=(vars20)(vars21)] (vars2){};
      \node[fit=(vars30)(vars31)] (vars3){};
      
      \draw [rectangle connector=1cm] (supervars) to  node[right] {} (vars0);
      \draw [rectangle connector=1cm] (supervars) to  node[right] {} (vars1);
      \draw [rectangle connector=1cm] (supervars) to  node[right] {} (vars2);
      \draw [rectangle connector=1cm] (supervars) to  node[right] {} (vars3);
      
      \node at ($(bn2s)!.5!(bn3s)$) {\vdots};
      \node at ($(vars2)!.5!(vars3)$) {\vdots};

    \end{tikzpicture}
   
    \caption{Graphical representation of the metamodel. This visualization assumes that the generator and the discriminator are codified within a single individual. If the DNNs were evolved separately, a single submodel per DNN per depth would be enough. The example BN graphs do not represent a realistic scenario in terms of number of nodes. $l$ and $m$ represent the maximum number of hidden layers in the discriminator and generator, respectively.}
    \label{fig:metamodel}
  \end{center}
\end{figure}
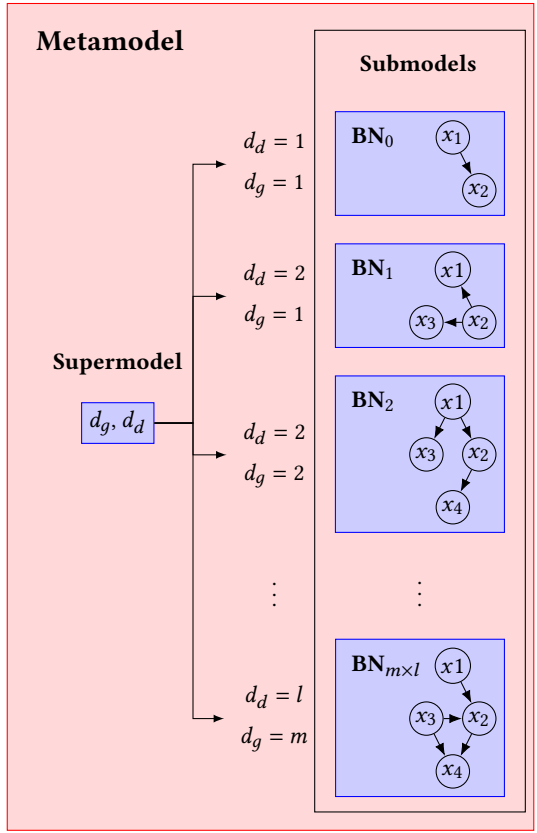

Most current NAS algorithms allow the development of DNNs of varying depth, and the more layers in a DNN, the longer the list of components required to characterize it. This leads to DNN characteristic vectors being dynamic, and dealing with different variables in a single dataset is not straightforward for a BN model.
Because of this, we have chosen to design the metamodel proposed in this work at two levels. The first level, the \textit{supermodel}, fixes the number of layers of the generator and the discriminator. Next, in the second level, there is one BN learned for each DNN depth. In cases when the GAN is considered as a sole entity, one \textit{submodel} is learned per each depth combination of the two networks. If, conversely, generators and discriminators are developed independently, a single BN per depth of each network is enough. Figure~\ref{fig:metamodel} shows a graphical representation of the metamodel.

In this work, after a preliminary experimentation, we decided to employ the ARACNE algorithm (an extension to the Chow-Liu algorithm \cite{chow_approximating_1968}) \cite{margolin_aracne_2006} to learn the structure of the BN. All the BN related implementation is based on the bnlearn R library \cite{scutari_bayesian_2014}.

\section{Main experiments} \label{sec:mainExp}

The goal of this experimental section is to test the validity of the proposal at different levels, namely, its capacity to distinguish \textit{good} from \textit{bad} structures, its ability to sample strong GAN structures, and the suitability of these as the starting point for a NE procedure.

This experimental section is based on the evolutionary runs provided by the authors of \cite{garciarena_analysis_2020} (Section~\ref{sec:rw} contains a short description of the evolutionary procedure). Overall, for this part, there are $8\text{ }functions\times 2\text{ }variable sizes\times 30\text{ }runs=480runs$. The functions from the suite defined in \cite{li_multiobjective_2008} being used in this work are $F1$, $F2$, $F3$, $F4$, $F5$, $F7$, $F8$, and $F9$. 

\subsection{GAN choice}

In order to nurture the metamodel with characteristics of GANs that posteriorly make the metamodel as useful as possible, we have defined three separate sets of GANs. The first set, \emph{First}, consists of the $n$ best performing individuals of each run. The second one, \emph{Second}, is composed of the $n$ second best individuals, that is, from the $n-th$ to the $2n-th$ in the ranking ordered by the chosen quality metric. Finally, the \emph{Random} set contains $n$ random individuals from each run. The \emph{First} set will be used to learn the metamodel, while the \emph{Second} set will help test its generalization capacity, and the \emph{Random} set acts as a control set, determining whether the metamodel is able to discriminate individuals with great potential from mediocre ones. In this particular case, we use $n=5$ individuals. 

In this approach, the two DNNs in a GAN are evolved together and they have a maximum of 11 hidden layers and a minimum of one, resulting in 100 possible combinations. We have observed that more than $75\%$ of the MLP-GANs in \emph{First} are combinations of generators of a depth up to 3 layers, and discriminators with no more hidden layers than 4. In order to reduce the complexity of the model built and its interpretability, we have decided to limit the GANs to these $3\times4=12$ of the total $10\times10=100$ depth variants.

The training frequency and the number of neurons in each layer are the only continuous variables describing a GAN. For the sake of simplicity, we have discretized them to 5 possible different values. A possible extension to this limitation is left as future work.

\subsection{Metamodel testing}

We first develop and validate our approach in the MLP-GAN\footnote{In the original contribution, tensorflow \cite{abadi_tensorflow_2016} was used to implement the experiments.} domain. Two different methods to determine the extent to which the proposed metamodel has been able to capture the dependencies between the components of the GANs have been defined. The first step is to test whether the metamodel is indeed able to \textit{recognize} top quality GANs, and once this is done, test whether it can \textit{produce} such generative models.

\subsection{GAN Likelihood} \label{sec:likelihood}

\begin{figure*}
	\begin{center}
			\centering
			\includegraphics[width=0.9\textwidth,trim={0 0 0 0},clip]{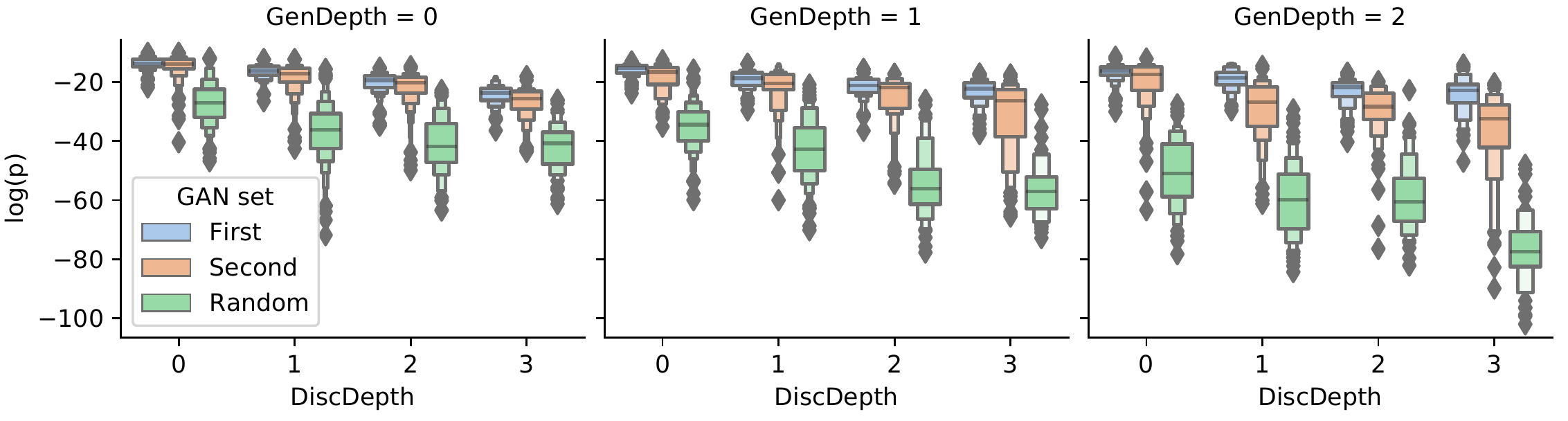}
			\caption{Probabilities (in logarithmic scale) provided by the metamodel to GANs present in the three defined sets.}
		\label{fig:probas}
	\end{center}
\end{figure*}

One efficient way of testing whether or not what the metamodel has learned from the provided data can be valid, is to compare the probability given by the metamodel to GANs from the \emph{First} (the set of GANs from which the metamodel is learned), \emph{Second}, and \emph{Random} sets. If the probability values provided by the metamodel to GANs from the \emph{First} and \emph{Second} set are similar, and the values between \emph{Second} and \emph{Random} are different, we can determine that the model is good at both generalizing to unseen \textit{good} structures and discerning \textit{bad} structures.

Figure~\ref{fig:probas} shows three boxplots (one per generator depth) in which the probabilities (in the y axis in logarithmic scale) assigned by the metamodel to the GANs of each set are displayed. The x axis represents the depth of the Discriminator. As can be seen, the metamodel grants similar probability values to GANs in \emph{First} and \emph{Second}, at the same time it is able to discern GANs from \emph{Random} by assigning them considerably lower probabilities.

The apparent differences between the probability sets are confirmed by the Kruskal-Wallis statistical test \cite{kruskal_use_1952}, which provided a maximum p-value of $\sim10^{-27}$ for the $4 disc. \times 3 gen. depths = 12$ different combinations tested. The Dunn statistical test \cite{dunn_multiple_1964} confirmed that all pair-wise comparisons were different, as the maximum p-value found for the $12 comb. \times 3 pairs = 36 pairs$ tested was $\sim0.04$. However, the differences on the p-values provided by the Dunn test varied vastly depending on the pairs being compared. Between the \emph{First} and \emph{Second} sets, the p-values ranged from $\sim0.04$ to $\sim3\times10^{-26}$, whereas comparing any of these two sets to \emph{Random} yields p-values between $\sim10^{-16}$ and $\sim10^{-161}$. This shows that the \textit{gap} between the probabilities of the \emph{First} and \emph{Second} set GANs is much narrower than the distance between any of these sets to the \emph{Random} set.

\subsection{Sample generation capacities}

To assess the sampling capacities of the proposed metamodel, we firstly define two variations of the \emph{First} and \emph{Random} sets. $\text{First}_\text{train}$ is a restricted version of \emph{First} which, instead of including structures from all $480$ runs, only $300$ are considered, those corresponding to 5 of the 8 total functions (F1, F2, F3, F4, and F7). $\text{Random}_\text{train}$ is populated similarly, but with $n$ random GANs instead of the best found ones. The selection of the functions in each set has been performed according to the function similarity criterion deduced in \cite{garciarena_analysis_2020}. Further inclusion of problem characterization into the metamodel to achieve even greater results is left as future work.

\begin{figure}
	\begin{center}
			\centering
			\includegraphics[width=0.47\textwidth,trim={0 0 0 0},clip]{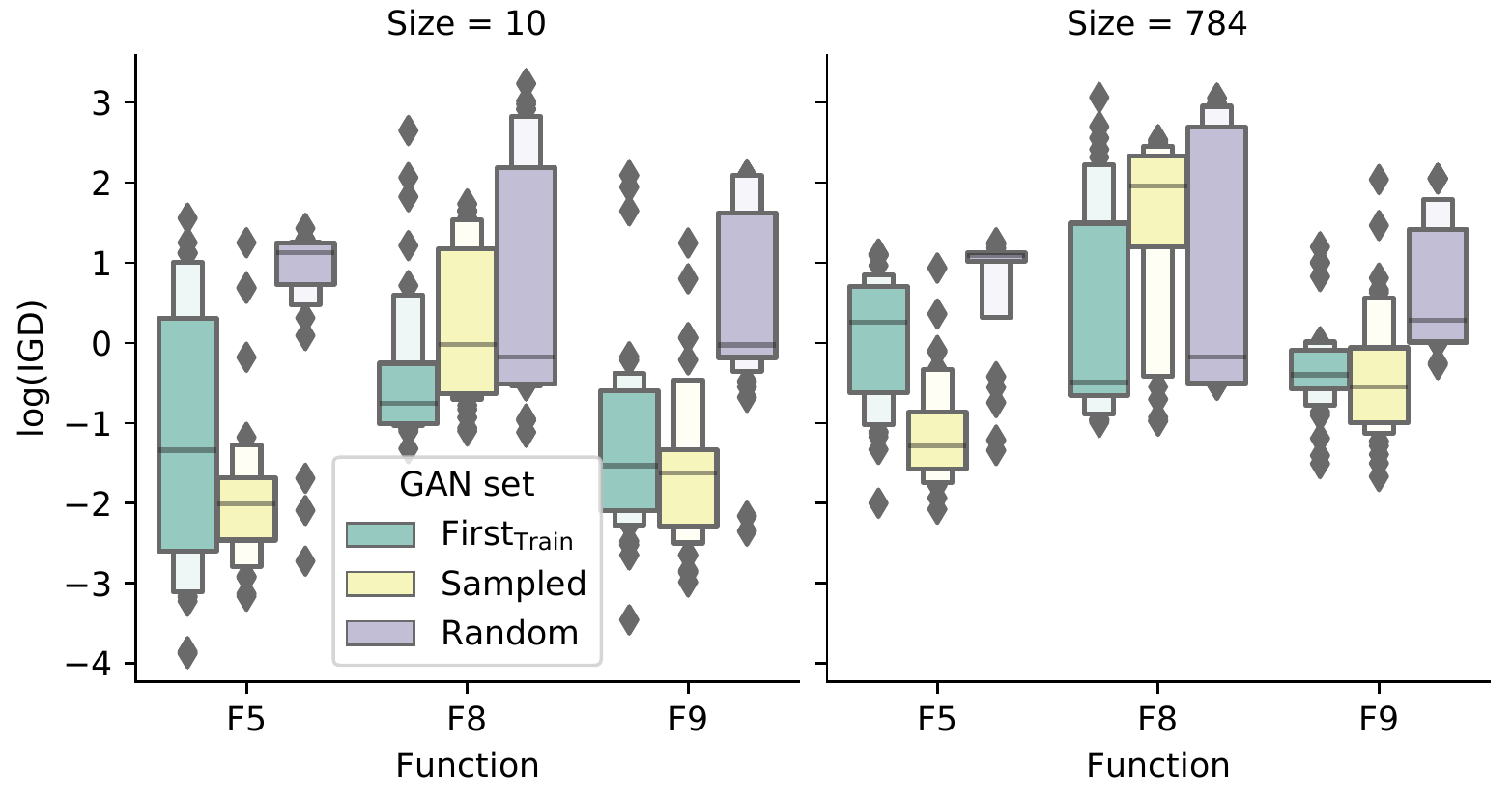}
			\caption{IGD values obtained by GANs in \emph{First}, GANs sampled from the metamodel, and GANs from \emph{Random}.}
		\label{fig:samples}
	\end{center}
\end{figure}

A new metamodel is learned using the reduced \emph{First} set, $\text{First}_\text{train}$, before being sampled $100$ times. This is achieved by using probabilistic logic sampling (PLS) \cite{henrion_propagating_1988}, a method that samples variables following their ancestral order.
These samples conform the Sampled set. Another $100$ GANs are randomly chosen from $\text{First}_\text{train}$ and $\text{Random}_\text{train}$. All these structures are trained to reproduce Pareto set approximation of the functions that have remained isolated from this procedure (F5, F8, and F9), for both variable sizes. The fitness function used to evolve the GANs, the IGD values between the samples obtained from the GANs and the real PF, obtained by these structures in the different problems are displayed in Figure~\ref{fig:samples}.

As can be observed in the figure, the results vary depending on the function on which the test is performed, and are consistent through problem size. Regarding F5, the functions for which low IGD values were recorded, the sampled GANs performed surprisingly well, clearly outperforming the random GANs, and arguably performing better than the GANs chosen from \emph{First}. This is not the case for the more difficult function F8, for which the best GANs yielded the best results, and the random GANs offered similar results to the sampled ones. Tests performed using F9 were not able to discern an approach that is better than the other between the best and the sampled GANs, while both of these sets clearly outperformed the random GANs. The Kruskal-Wallis test reported significant differences in all 6 trios of IGDs. The Dunn test produced p-values under $3\times 10^{-7}$ for all pairwise comparisons for F5 except for when comparing the \emph{First} and \emph{Sampled} GANs with 10 variables, where the p-value was $0.12$. This means that, in the worst of cases, the sampled GANs performed as well as the best ones, if not better. The random ones did not offer competitive results. For F8, comparisons between random and sampled GANs yielded p-values of $0.36$ and $0.96$ for 10 and 784 variables respectively, whereas all other comparisons resulted in p-values smaller than $0.002$. In this case, sampled GANs did not offer better results than random ones, and both these sets of GANs were one step below the best GANs. Finally, for F9, the comparison between best and sampled GANs showed no clear differences (p-values of $\sim0.2$), while the other comparisons resulted in p-values below $5\times 10^{-8}$. Similar conclusions to the ones extracted from F5 can be deduced in this case.

As a general conclusion of this analysis, we could say that, in terms of the IGD, (the metric that was minimized during the evolution of the GANs in \emph{First}) the sampled GANs sometimes offer worse results than random GANs would (F8 with 784 variables). However, in most cases, sampled GANs offer a similar performance to GANs in \emph{First} (F9, or F8 with 10 variables), if not better (F5). Interestingly, a small selection of random structures were able to provide a similar performance to the more sophisticated ones, which indicates that efficient learning procedures can also be carried out even when the GAN structure is constructed without any knowledge.

\subsection{Metamodel for improving GAN structural searches}

\begin{figure}
	\begin{center}
			\centering
			\includegraphics[width=0.43\textwidth,trim={0 0 0 0},clip]{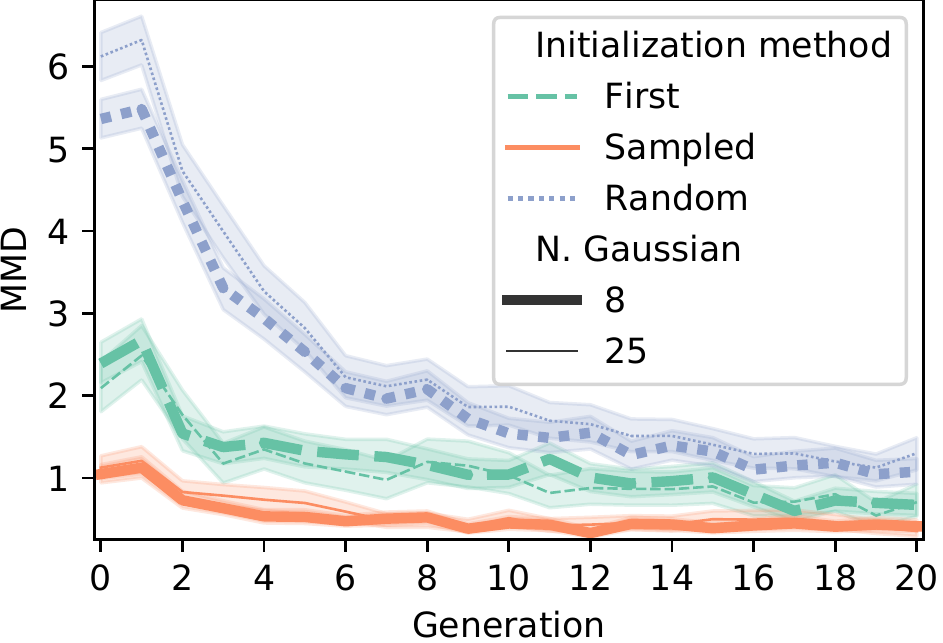}
			\caption{Best MMD value (y axis) corresponding to the samples generated by GANs at different generations (x axis) during the evolutionary procedures. The solid lines represent the mean of the $20individuals\times generation*30=60$ computed MMDs at each generation, whereas the translucent bands show the $95\%$ confidence interval of all these runs. Runs for the 2 variations of the problem and three NE initialization methods are displayed.}
		\label{fig:MMDevo}
	\end{center}
\end{figure}

In the previous section, the ability of generating capable GANs from the proposed metamodel was shown. Now, we put that capacity to the test, by proposing different ways of initializing individuals in other NE algorithms applied to a completely different problem. For this purpose, we design a GAN structural search for the 2D 8 and 25-Gaussian approximation problem \cite{metz_unrolled_2016}, which is specifically formulated to expose mode collapsing GANs, one of the most significant flaws of these generative models. Thus, we set a reduced version of the evolutionary process proposed in \cite{garciarena_analysis_2020}, consisting of populations composed of 20 individuals and 20 generations. Three versions of the NE algorithms are run, the difference between them being the way the initial population is created. Sets similar to those defined in the previous section are defined (\emph{First}, \emph{Sampled}, and \emph{Random}), only this time no runs are left out when learning the metamodel (previously the runs corresponding to three functions were excluded). Each variant of the NE algorithm used GANs from one of the three sets to fill the initial population. 30 runs of each evolutionary run were performed. Figure~\ref{fig:MMDevo} shows the per generation evolution of the best GAN in terms of Maximum Mean Discrepancy (MMD) \citep{gretton_kernel_2012}, the fitness function used to evaluate the quality of the generations of a GAN (the second objective of the bi-objective evolutionary process being the minimization of the elapsed time during training and sampling the GANs). The MMD improvement for $2 dimensions \times 3 initializations=6$ run types is shown.

As can be seen in the figure, the evolutionary runs with the non-random initialization have a large advantage in the initial stages of the evolution, as could have been expected. Even though it is true that the randomly initialized runs experience larger gains over the course of the procedure compared to the other two algorithms, it is not enough to produce competitive results. Something similar can be said about the runs initialized by the GANs in \emph{First} and \emph{Sampled}, as the GANs in the initial population of the \emph{Sampled} run are not outperformed by any of the found architectures in the runs initialized with GANs in \emph{First}.

This figure clearly shows the advantages of using the metamodel for initializing the first population of a NE procedure. Instead of using GANs which were specifically evolved for diverse tasks, learning the more general characteristics of GANs evolved across different runs for different problems provides the metamodel-aided approach with the necessary generalization capacities to balance the available information exploitation. This could be due to the structures in \emph{First} \textit{overfitting} the problem they were evolved for. Because the metamodel cannot learn to generate GANs from \emph{First} exactly as they are, and it has to focus on \textit{only} capturing the dependencies within the structures that make them perform as they do, the GANs that it posteriorly produces are able to generalize better. This generalization capacity is a desired characteristic in many scenarios, including the one described in this experimentation.

\section{Application to CNN-GANs} \label{sec:secondaryExp}

In the previous Section~\ref{sec:likelihood}, the metamodel has shown its capacity to \textit{classify} GAN structures according to how likely they are to offer a strong performance. NAS algorithms would greatly benefit by the usage of a model that could indicate the structures that are likely to perform poorly, avoiding unnecessary evaluations, and thus speeding the whole process up. GANs composed of CNNs are specially costly to be evaluated, and we therefore consider this special case of NE as a suitable field to test the capacity of the model to make processes more agile.

Firstly, COEGAN is run for a single database, Fashion MNIST \cite{xiao_fashion-mnist:_2017}, 20 times. Because the two DNNs are evolved separately, and they are formed of no more than 6 layers, only 12 BN submodels have to be learned. \footnote{All the DNNs of this section have been implemented using the PyTorch library \cite{paszke_pytorch_2019}.}
Next, a \emph{First} set of GAN was created, and a metamodel similar to that introduced in Section~\ref{sec:design} was learned. Because of the reduced number of runs, we use $n=20$ so that the metamodel has enough examples to learn from.

Finally, 30 different runs of a hill climbing (HC) algorithm are executed looking for GAN structures which can accurately reproduce images similar to the digits available in the MNIST dataset \cite{lecun_mnist_2010}. Each HC run is awarded a limit of 100 evaluations, and the difference between the two variants designed for this experiment resides in the usage of the metamodel to guide the direction in which the algorithm will move. While the first variant will simply randomly generate a neighbor to the current GAN being evaluated, the second one generates as many neighbors as possible (in both cases generated using the mutation operators defined for COEGAN), check which one is the most likely to make the largest immediate improvement based on the probabilities assigned by the metamodel, and choose it as a candidate.

\begin{figure}
	\begin{center}
			\centering
			\includegraphics[width=0.47\textwidth,trim={0 0 0 0},clip]{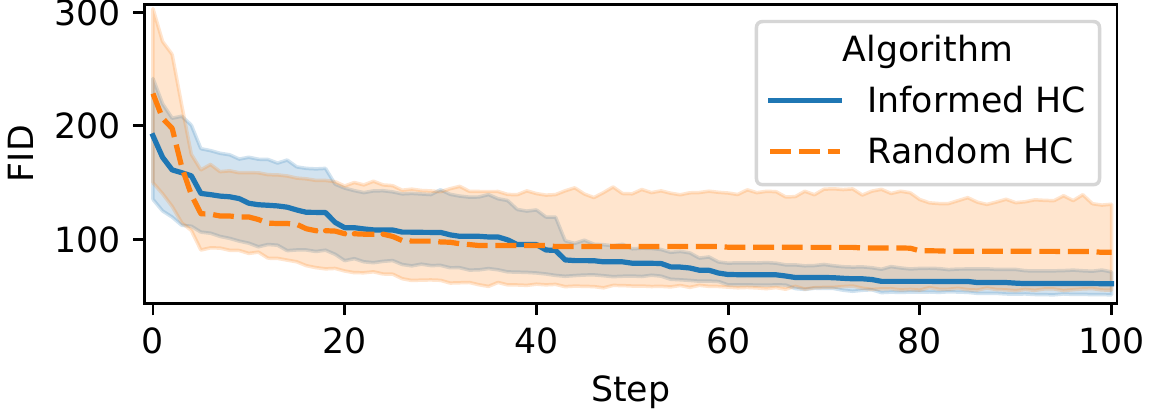}
			\caption{Evolution of the FID (in the y axis) of the best CNN-GAN structure found at each step (x axis). The solid line represents the median of all 30 runs (10 right now, the rest are being executed, possibly done by Sunday). The translucent bands represent the 95\% confidence interval. The orange, discontinued line represents runs performed with the random hill climbing, whereas the blue continuous line does so for the hill climbing guided by the metamodel.}
		\label{fig:FIDEvo}
	\end{center}
\end{figure}

\begin{figure}
	\begin{center}
			\centering
			\includegraphics[width=0.28\textwidth,trim={0 0 0 0},clip]{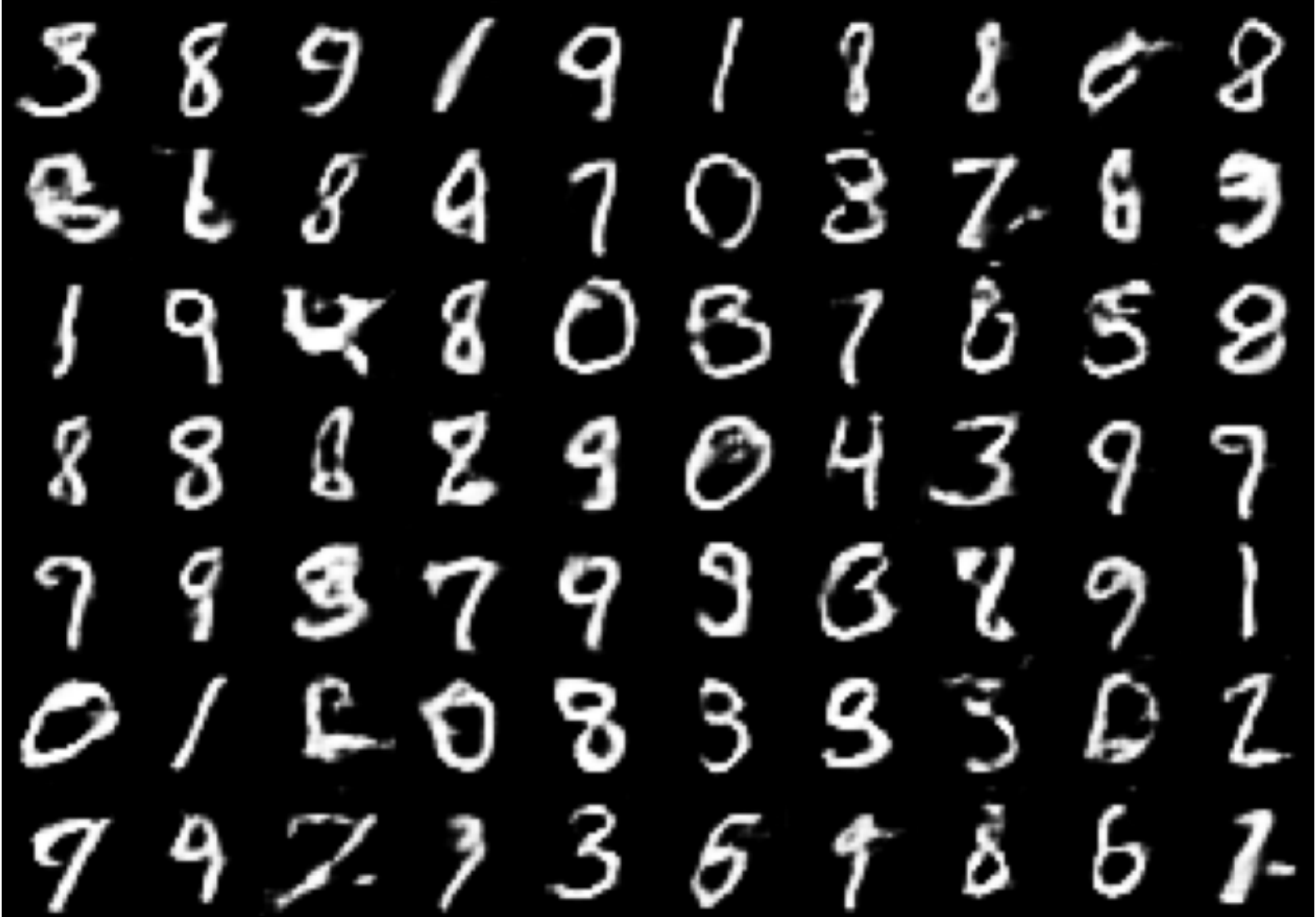}
			\caption{Examples of digits generated by the best structure found by the informed hill climbing runs.}
		\label{fig:MNIST}
	\end{center}
\end{figure}

Figure~\ref{fig:FIDEvo} shows, for each step in the HC procedure (x axis), the best found FID value (y axis). Similarly to Figure~\ref{fig:MMDevo}, the solid lines show the median run, and the translucent bands show the 95\% confidence interval. The figure shows that, during the first 40 steps of the search, both HC procedures show similar behaviours, with a slight advantage for the random search. In the second part of the search, however, only the guided greedy algorithm is capable of showing steady improvement, whereas most of the random HC runs get stagnant from the 40-th step onward. This shows the benefit of a guidance during this search, as it avoids getting stuck local minima.
Some samples extracted from the best GAN (in terms of FID) are shown in Figure~\ref{fig:MNIST}.

\section{Conclusions and future work} \label{sec:conclusions}

In this paper, we have proposed a methodology that makes the most out of the computational effort performed during a neural architecture search. We tested the methodology on two neuroevolutionary algorithms that search for GAN structures whose success has been reported in the literature. This proposal consists of modeling the best architectures found in optimization procedures with a metamodel based on Bayesian networks, with two main goals. 

Firstly, by training the metamodel with the best individuals found during previous evolutionary procedures, the model is able to recognize networks which belong to the distribution of GANs likely to deliver a top performance. This way, posterior neural architecture searches could lighten the computational effort required to be carried out by ignoring individuals unlikely to offer a top performance. 

Secondly, the metamodel is able to sample GAN specifications which are likely to offer a good performance without any structural optimization being performed on them. 
 
We first successfully tested the two mentioned capacities of the proposed metamodel isolated from a structural optimization process, and posteriorly put them into practice using two proven neural architecture search procedures. In the first successful test, the metamodel was employed on the same (or similar) problem, and the two following tests were carried out in more diverse scenarios, resulting in a strong validation of the proposed methodology.

In the first case, the metamodel learned with MLP-GANs evolved for the PS approximation is used to initialize a population. This population is used to conduct a secondary neuroevolutionary process which seeks GAN for the 2-D Gaussian mixture approximation problem. Secondly, we also tested the capacity of the model to speed up a neural architecture optimization algorithm. This time, a metamodel learned from CNN-GANs evolved for reproducing samples from the Fashion-MNIST dataset is used to guide a local search for GANs whose goal is to reproduce MNIST data. The results show how the extra information provided by the metamodel helps the search algorithm to keep finding increasingly better structures where a random version of the same algorithm could not.

The results reported in this work suggest a large potential for these metamodels, as a version learned from a rather modest variety of data was able to produce interesting results. Taking into account the amount of neuroevolutionary procedures being carried out in the world, the generalization capacity of a more complex version of the model proposed in this paper learned from more sophisticated and abundant data could reach heights that would enable them to be applied to any neural architecture searches.

\begin{acks}
This work has received support from the TIN2016-78365-R (Spanish Ministry of Economy, Industry and Competitiveness), PID2019-104966GB-I00 (Spanish Ministry of Science and Innovation), IT-1244-19 (Basque Government), KK-2020/00049 (part of the project 3KIA, funded by the SPRI-Basque Government through Elkartek) programs, and by Portuguese national funds through the FCT - Foundation for Science and Technology, I.P., through the project CISUC - UID/CEC/00326/2020. Unai Garciarena holds apredoctoral grant (PIF16/238) by the University of the Basque Country.
\end{acks}

\bibliographystyle{ACM-Reference-Format}
\bibliography{references} 

\end{document}